\documentclass[conference]{IEEEtran}
\IEEEoverridecommandlockouts

\usepackage{cite}
\usepackage{amsmath,amssymb,amsfonts,bbm}
\usepackage{float}
\usepackage{graphicx}
\usepackage{textcomp}
\usepackage{xcolor}
\usepackage{bm}
\usepackage{subfigure} 
\usepackage{romannum}
\usepackage{xparse}
\usepackage{algorithm, algpseudocode}

\def\BibTeX{{\rm B\kern-.05em{\sc i\kern-.025em b}\kern-.08em
    T\kern-.1667em\lower.7ex\hbox{E}\kern-.125emX}}
\begin{document}

\title{Synthetic Power Analyses: Empirical Evaluation and Application to Cognitive Neuroimaging}

\author{
\IEEEauthorblockN{Peiye Zhuang}
\IEEEauthorblockA{\textit{Dept. of Comupter Science,} \protect\\
\textit{University of Illinois} \protect\\ 
\textit{at Urbana-Champaign} \protect\\ 
peiye@illinois.edu}\\
\and
\IEEEauthorblockN{Bliss Chapman}
\IEEEauthorblockA{
\textit{Apple}\\
bliss.chapman@gmail.com}
\and
\IEEEauthorblockN{Ran Li}
\IEEEauthorblockA{ \textit{Google} \\
ryannli1129@gmail.com}
\and
\IEEEauthorblockN{Sanmi Koyejo}
\IEEEauthorblockA{\textit{Dept. of Comupter Science} \protect\\
\textit{\& Beckman Institute,} \protect\\
\textit{University of Illinois} \protect\\ 
\textit{at Urbana-Champaign} \protect\\
sanmi@illinois.edu}
}

\newcommand{\norm}[1]{\left\lVert#1\right\rVert}
\newcommand{\B}[1]{\bm{#1}} 
\newcommand{\ind}[1]{\left\llbracket#1\right\rrbracket}
\algnewcommand\Input{\item[\textbf{Input:}]}%
\algnewcommand\Output{\item[\textbf{Output:}]}
\algnewcommand\RETURN{\item[\textbf{Return:}]}
\maketitle

\begin{abstract}
 In the experimental sciences, statistical power analyses are often used before data collection to determine the required sample size. However, traditional power analyses can be costly when data are difficult or expensive to collect. We propose synthetic power analyses; a framework for estimating statistical power at various sample sizes, and empirically explore the performance of synthetic power analysis for sample size selection in cognitive neuroscience experiments. To this end, brain imaging data is synthesized using an implicit generative model conditioned on observed cognitive processes. Further, we propose a simple procedure to modify the statistical tests which result in conservative statistics. Our empirical results suggest that synthetic power analysis could be a low-cost alternative to pilot data collection when the proposed experiments share cognitive processes with previously conducted experiments.
\end{abstract}
\begin{IEEEkeywords}
  fMRI, GANs, power analyses
\end{IEEEkeywords}
\vspace{-0.2cm}
\section{Introduction}
\vspace{-0.2cm}
\label{introduction}
Cognitive neuroscience studies how the brain produces intelligent behavior. One of the most common measurement tools for brain activity is functional magnetic resonance imaging (fMRI), used to gain insight into how patterns in brain activity correspond to cognitive functions. The use of imaging tools to understand cognition is known as cognitive neuroimaging. This paper proposes generative modeling techniques that provide a promising methodological approach to improve statistical analysis in cognitive neuroimaging.

Improvements to statistical analysis are particularly timely due to an increasing focus on the quality of data-dependent analysis in science and medicine, probably most publicly in the psychological sciences~\cite{open2015estimating, pashler2012editors} One of the primary quantities of interest in statistical analyses of scientific data is statistical power; the probability that a test rejects the null hypothesis given that the alternative hypothesis is correct. Lower power reflects a low probability that the test correctly detects a statistically significant effect. Neuroscientists typically aim for 80\% power~\cite{mumford2012power}.
 However, in recent years, the statistical power of published neuroscience studies has been demonstrated to be much lower than initially claimed, especially in published fMRI experiments~\cite{durnez2016power}.

Prospective power analyses promise an optimal experiment design that maximizes statistical power and reduces the waste of resources on futile experiments or additional subjects. In traditional power analysis, real data is collected and used to determine the sample size needed to achieve the desired experimental power. Unfortunately, an MRI machine with resolution at the level required for neuroscience research costs between 0.5 and 3 million dollars~\cite{glover_2014}, and the scanning can cost about \$500+ an hour~\cite{fmriarterialspinlabeling}. Thus, resources are not always available to run studies that could supply data for power analysis. This barrier lowers the quality of proposed research experiments and slows overall progress in neuroscience.

One approach to circumvent the data problem is to use stand-in data from other research groups for power analysis. However, even when the borrowed data is a good substitute for the proposed experiment, this approach requires planning well in advance and slows turnaround times. More importantly, finding stand-in data can be challenging. An alternative approach is to determine effect sizes and variance estimates from previously published studies. Again, even when one can find a sufficiently similar study, the experiment can suffer from underpowered estimates because published articles are biased towards reporting significant effects~\cite{open2015estimating, pashler2012editors}.

We propose the use of modern generative neural network models to synthesize diverse, high-quality brain images. Specifically, we use the three-dimensional Improved Conditional Wasserstein Generative Adversarial Network (ICW-GANs) model~\cite{zhuang2018hallucinating}. The observation motivates our use of this generative approach to produce unlimited synthetic data, which, in turn, promises informed power analyses with minimal cost overhead. In particular, our model can synthesize cognitive processes corresponding to labels that are not present in the original training set, thus enabling the direct application of synthetic fMRI data to real-world power analyses for new experiments. 

The primary aim of this manuscript is to empirically evaluate the performance of synthetic power analyses through simulated data and neuroimaging experiments. We compare the power analysis results of synthetic to real data using both the classic two-sample t-test and the non-parametric maximum mean discrepancy (MMD) test~\cite{gretton2012kernel}.
Our experiments demonstrate that the synthetic data has similar distributional characteristics as real data, and that power analysis results calculated with synthetic data are similar to those calculated with real data. Taken together, our results suggest that synthetic data can be a reliable and low-cost substitute for real data used in cognitive neuroimaging experiment design.

\vspace{-0.3cm}
\section{Related Work}
\vspace{-0.2cm}
\label{related_work}



\textbf{Availability of brain imaging Data.}
The lack of sufficient brain imaging data may stifle the progress in computational cognitive neuroimaging research~\cite{poldrack2014making}.  This issue is more evident in {\em decoding} studies predicting cognitive and behavioral outcomes of brain imaging experiments, particularly when one would like to use machine learning methods ~\cite{varoquaux2014machine}. As a result, there is a dearth of research on sophisticated predictive models for high-dimensional brain images~\cite{cox2003functional, pereira2009machine, nathawani2016neuroscience}.

\textbf{Deep learning in Neuroscience.}
Neural networks have been used extensively for classifying brain imaging data. For instance, deep networks have been used to extract features of fMRI brain images to classify brain states \cite{firat2014deep, koyamada2015deep}, and 2D and 3D neural networks to classify fMRI brain data~\cite{nathawani2016neuroscience}. fMRI data of video stimuli have been decoded and classified data into visual categories \cite{svanera2017deep}. Similarly, \cite{nathawani2016neuroscience} extracted features from 4D fMRI data and used deep learning methods for discrimination of cognitive processes.

\textbf{Generative Adversarial Networks (GANs).}
Generative Adversarial Networks (GANs) \cite{goodfellow2014generative} are a promising recent development in generative modeling. GANs have demonstrated a capacity to capture complex data distributions through a non-cooperative two-player game formulation. 
In the GAN framework, a generator $G$ transforms samples from a simple distribution into complex high-dimensional data. At the same time, a critic $D$ attempts to distinguish between the synthetic and real data. 
Let $x$ represent the data with empirical data sensity $p_{data}(x)$. The variable $z$ represents the low-dimensional latent variable with a simple distribution $p(z)$ such as $\mathcal{N}(0,I)$. The generator and the critic optimize value function $V(G, D)$ through a two-player minimax game:
\vspace{-0.1cm}
\begin{equation}
\begin{aligned}
\min_{G} \max_{D} V(D, G) = & \mathbb{E}_{x \sim p_{data}(x)}[\log(D(x))]  \nonumber \\
& + \mathbb{E}_{z \sim p_z(z)}[\log(1 - D(G(z)))].
\end{aligned}
\end{equation}

In the conditional GAN \cite{mirza2014conditional}, the model is provided additional information at 
training time about class labels or features. 
3D-GANs \cite{
mirza2014conditional} extend GANs to 3D object generation through the use of three-dimensional
convolutions.
However, the classical training procedure is brittle without other stabilization techniques.
The conditional information can be manipulated to allow for finer-grained control over data generation. The Wasserstein GAN (WGAN) \cite{
arjovsky2017wasserstein} uses Wasserstein-1 distance as a loss metric that improves the stability of the training procedure. Improved Wasserstein GAN (WGAN-gp) \cite{gulrajani2017improved} further extends these stability improvements by introducing a gradient penalty term in the critic network that replaces weight clipping and enforces a Lipschitz constraint.  \cite{zhuang2018hallucinating} first proposed an Improved Conditional Wasserstein 
Generative Adversarial Networks (ICW-GAN) to the synthesis of fMRI brain imaging data. The objective function of a 3D Conditional Wasserstein ICW-fMRI-GAN is as follows:

\begin{equation}
\begin{aligned}
L = &\mathbb{E}_{\B{z} \sim P_{\B{z}}(z)} [1 - D(G(\B{z}|\B{y}))] \\
& - \mathbb{E}_{\B{x} \sim P_{data}(x)} [D(\B{x}|\B{y})] \\
& + \lambda \mathbb{E}_{\hat{\B{x}} \sim P_{\hat{\B{x}}}(\hat{x})}[(\norm{\nabla_{\hat{\B{x}}}D(\hat{\B{x}}|\B{y})}_2 - 1)^2],
\end{aligned}
\label{eq:icw-gan}
\end{equation}

where $\B{y}$ denotes the volume labels, $\hat{\B{x}} \leftarrow \epsilon \B{x} + (1 - \epsilon)G
(\B{z})$, $\epsilon \sim U(0, 1)$ and $\lambda$ is a gradient penalty coefficient.

\textbf{Two-sample test.} In statistical testing, a two-sample test is used to deduce if two samples are generated from the same distribution or two separate distributions.
We consider two main types of two-sample tests, as outlined in the following.

i) A parametric test based on explicit distributional assumptions e.g., t-test\cite{ttest1, ttest2, ttest3, ttest4} assumes that the two distributions are both Gaussian distributions.

ii) A non-parametric test that employs weaker assumptions, while yielding effective tests. One example is the maximum mean discrepancy (MMD) test \cite{mmd1, mmd2, mmd3}. Suppose there are two distributions  ${\displaystyle P(X)}$ and ${\displaystyle Q(Y)}$, the MMD between them can be formulated as follows: 
\vspace{-0.1cm}
$${\displaystyle {\text{MMD}}(P,Q)=\sup _{f \in \mathcal{F}}\left(\mathbb {E} _{X}[f(X)]-\mathbb {E} _{Y}[f(Y)]\right)},$$
where $f$ is a function in a function set $\mathcal{F}$. Usually, a unit ball in a reproducing kernel Hilbert space (RKHS)\cite{smola2007hilbert} is used as the MMD function classes $\mathcal{F}$. Classical MMD methods compute $L_2$ distance between two distributions in the kernel space. Recent work \cite{scetbon2019comparing} has shown that replacing the distance with the $L_1$ distance can improve test power. 
\section{Methods}
\label{method}
\vspace{-0.2cm}
\subsection{ICW-GANs}
\vspace{-0.1cm}
\label{sec:icwgans}
We use an ICW-GAN, a class-conditional model proposed in \cite{zhuang2018hallucinating} to synthesize fMRI data conditioned on selected labels. The objective function of an ICW-GAN is shown in Eq.~\ref{eq:icw-gan}.
With regard to its architecture, the generator consists of a four-layer fully-connected convolutional layers, batch normalization, ReLU layers added in between, and a sigmoid layer at the end. We use 3-dimensional convolutions which enable modeling of the spatial relations of high-dimensional brain images. The critic has a similar architecture as the generator except for the final layer that uses a linear activation.
We train for 15,000 iterations with batches of size 50, an input noise vector of length 128, and a gradient penalty $\lambda=1$. Both the critic and the generator are optimized with the RMSprop optimizer~\cite{rmsprop} with a learning rate of $1e-3$ and are updated once per step.
\vspace{-0.1cm}
\subsection{Synthetic Power Analyses}
In a traditional power analysis, real data is collected and used to determine the sample size needed to achieve a desired experimental power. Let $p$ be the p-value returned by a statistical test; denoting the probability of obtaining a result equal to or more extreme than what is observed under the assumption of no effect or no difference (null hypothesis). Let $\alpha$ be the pre-determined test threshold; denoting the rate of falsely rejecting the null hypothesis (type \Romannum{1} error) threshold (often set to $.1$ in neuroimaging studies). The power is denoted by $\gamma^*$, given by:
\vspace{-0.3cm}
$$\gamma^* = \mathbb{E}[ \mathbbm{1}{\{p < \alpha\}}].$$
\vspace{-0.1cm}
In our simulated experiments, we mimic a power analysis with both real and synthetic data by considering samples from the underlying distributions $D_1$ and $D_2$. The experimental procedure is outlined in Algorithm~\ref{alg:1}.
\vspace{-0.1cm}
\begin{algorithm}
	\caption{Power Analyses for two-sample test}
	\label{alg:1}
	\begin{algorithmic}[1]
		\Input{
		 two data distributions $D_1$ and $D_2$; 
		the number of samples 
		$n \in 	\{N_{start}, ..., N_{end} \}$; 
        the number of trials $K$.}
        \Output {$\gamma^{* (N_{start}, ..., N_{end})}$} 
		\State Train two generative models to recover estimates $\hat{D_1}$ and $\hat{D_2}$ of the underlying data distributions $D_1$ and $D_2$.
		
		\For{sample size $n = N_{start}, ..., N_{end}$}
		    \For{trial $k = 1, ..., K$}
		        \State Uniformly sample replicates of size $n$ from the real distributions $D_1$ and $D_2$ and the synthetic distributions $\hat{D_1}$ and $\hat{D_2}$.
                \State Compute statistics for tests distinguishing between the real and synthetic bootstrap replicates and calculate the $p$-values as $p^{real}_{n,k}$ and $p^{syn}_{n,k}$.
                \State Estimate the power $\gamma^{*(n)}_{real}$ of the test distinguishing between $D_1$ and $D_2$ and $\gamma^{*(n)}_{syn}$ of the test distinguishing between $\hat{D_1}$ and $\hat{D_2}$ with sample size $n$ by $\gamma^{*(n)} = \frac{1}{K} \sum_k \mathbbm{1}{\{p_{n, k} < \alpha\}}$.
            \EndFor
        \EndFor
        \RETURN $\gamma^{* (N_{start}, ..., N_{end})}$ 
	\end{algorithmic}  
\end{algorithm}
\vspace{-0.2cm}
  

For simulated data, we consider various approaches to sampling and testing. The first is direct sampling from the original distributions, which we denote as \textit{resampling}. Clearly, this is only possible with simulated data. We also consider \textit{ boostrap resampling}~\cite{efron1994introduction}. Here, we sample from the true distributions once. Then, during each iteration, we need only sample a subset from it. Finally, given the initial sample, we consider fitting the generative model, followed by standard \textit{resampling} from the synthetic distribution as a substitute for the real data.

\vspace{-0.3cm}
\subsection{Cognitive Neuroimaging Data}
\label{subsec:neuro}
The fMRI ICW-GAN introduced in Sec.~\ref{sec:icwgans} can synthesize fMRI data conditioned on cognitive process tag combinations that do not exist in the original dataset. Thus we can formulate synthetic power analyses with data that has never been collected before. To validate this approach, we provide details on our experimental procedure in Algorithm~\ref{alg:2}.

\begin{algorithm}
	\caption{Power Analyses for fMRI data}
	\label{alg:2}
	\begin{algorithmic}[1]
		\Input{
		 multi-tag fMRI dataset $D$; 
		the number of samples 
		$n \in 	\{N_{start}, ..., N_{end} \}$; 
        the number of trails $K$}
        \Output {$\gamma^{* (N_{start}, ..., N_{end})}$} 
        \State Train a generative model on $D$ to recover an estimate $\hat{D}$ of the underlying data distribution. 
		
		\State Split the dataset $D$ into two sub-samples $D=(D_0, D_1)$ where $D_1$ only contains samples with tag $x$ and $D_0$ contains the contrast tag (or other images).
		
		\For{sample size $n = N_{start}, ..., N_{end}$}
		    \For{trial $k = 1, ..., K$}
		        \State  Draw samples from the generative model. $\hat{D}_1$ are synthetic sampled conditioned on tag $x$, and $\hat{D}_0$ are synthetic examples \textbf{without} tag $x$.
		        \vspace{0.03cm}
		        
		        \State Conduct a statistical test between real data $D_0$ and $D_1$ to obtain the $p$-value as $p^{real}_{n,k}$. Conduct a statistical test between synthetic datasets $\hat{D}_0$ and $\hat{D}_1$to obtain the $p$-value as $p^{syn}_{n,k}$.
		        \vspace{0.0cm}
		        
                \State Estimate the power $\gamma^{*(n)}_{real}$ of the test distinguishing between $D_0$ and $D_1$ and the power $\gamma^{*(n)}_{syn}$ of the test distinguishing between $\hat{D_0}$ and $\hat{D_1}$ with sample size $n$ by $\gamma^{*(n)} = \frac{1}{K} \sum_k \mathbbm{1}{\{p_{n, k} < \alpha\}}$.
            \EndFor
        \EndFor
        \RETURN $\gamma^{* (N_{start}, ..., N_{end})}$ 
	\end{algorithmic}  
\end{algorithm}




For the fMRI data whose true distribution is unknown, we apply the  ``bootstrap sampling" method for real fMRI data. Similar to the settings in  the simulated data, the ``resampling" strategy is used for synthetic data.
As before, if at every sample size the power of a test computed between $\hat{D_0}$  and $\hat{D_1}$ is similar to the power of a test computed between $D_0$ and $D_1$, then using synthetic data as a stand-in for real data will yield a similar sample size estimation to what we would have achieved with real data.

\vspace{-0.3cm}
\section{Experiment}
\label{exp}
\vspace{-0.1cm}
\begin{figure}[t]
\begin{minipage}[b]{0.1\linewidth}
\centering
\footnotesize ``Non-visual"
\end{minipage}
\begin{minipage}[c]{0.95\linewidth}
\centering
\includegraphics[width=0.45\linewidth]{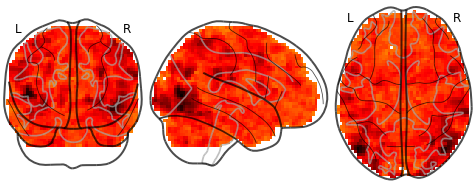}
\includegraphics[width=0.45\linewidth]{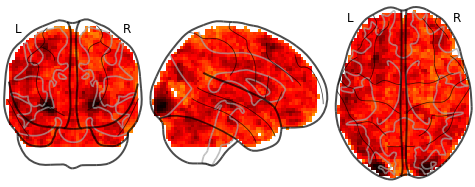}
\end{minipage}

\begin{minipage}[b]{0.1\linewidth}
\centering
\footnotesize ``Visual"
\end{minipage}
\begin{minipage}[c]{0.95\linewidth}
\centering
\includegraphics[width=0.45\linewidth]{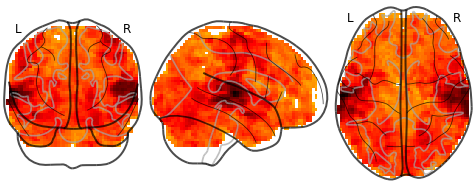}
\includegraphics[width=0.45\linewidth]{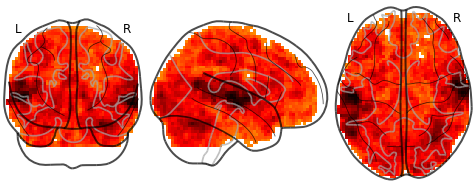}
\end{minipage}

\begin{minipage}[b]{0.1\linewidth}
\centering
\footnotesize ``Non-auditory"
\end{minipage}
\begin{minipage}[c]{0.95\linewidth}
\centering
\includegraphics[width=0.45\linewidth]{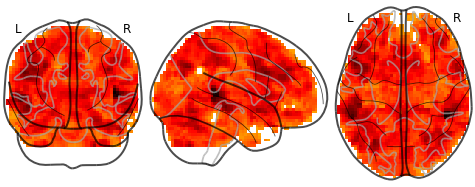}
\includegraphics[width=0.45\linewidth]{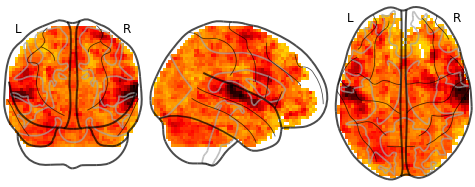}
\end{minipage}

\begin{minipage}[b]{0.1\linewidth}
\footnotesize ``Auditory"
\end{minipage}
\begin{minipage}[c]{0.95\linewidth}
\centering
\includegraphics[width=0.45\linewidth]{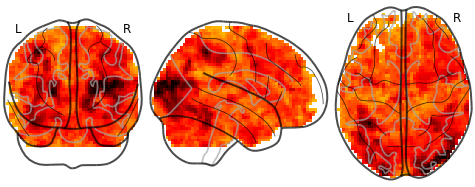}
\includegraphics[width=0.45\linewidth]{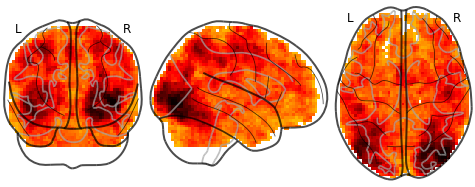}
\end{minipage}
\begin{minipage}[b]{\linewidth}
\centering
\hspace{2cm}
\footnotesize (a) Real images 
\hspace{1.2cm}
\footnotesize (b) Synthetic images
\end{minipage}
\caption{Visualization of real images (column (a)) and synthetic images (column (b)). Conditional labels are also shown.}
\label{fig:syn}
\vspace{-0.5cm}
\end{figure}

Next, we present our qualitative and quantitative evaluation of the proposed synthetic power analyses on simulated data in Sec.~\ref{exp:sim} and on fMRI data in Sec.~\ref{exp:fmri}. We  compare  the  power  analysis  results  between  synthetic  and  real  data  using  the  classical two sample t-test and the  non-parametric maximum mean discrepancy (MMD) test with an L1 norm as the measure of distance~\cite{scetbon2019comparing}.

For the fMRI brain images, we first present detailed qualitative results via examples of generated 3D volumes using the ICW-GAN. Then, we demonstrate prospective power analysis results for neuroimaging data.
\vspace{-0.2cm}
\subsection{Synthetic power on simulated data}
\label{exp:sim}
We employ two multivariate Gaussian distributions, $D_1 \sim \mathcal{N} (\mu = \textbf{0}, \Sigma = \mathbb{I})$  and $D_2 \sim \mathcal{N} (\mu = \textbf{0.3}, \Sigma = \mathbb{I})$ with a dimension of 10. We vary the sample sizes from 0 to 500 with a step size of 20, i.e., $N_{start}=0$ and $N_{end}=500$. For every sample size, we repeat the procedure 50 times to compute the power, i.e., $K=50$. We use two naive GAN models to generate examples for the two distrubutions of the simulated data. The structure of the generator and the critic is a  standard multilayer perceptron (MLP) network with 3 layers and ReLU activation between each layer. The input of the GAN are sampled data points from $D_1$ and $D_2$. We train the GAN models for 3,000 iterations with a batch size of 300. We visualize the loss curves of the GANs and synthetic data in order to guarantee that the GANs are well-trained. 
We compute power using a t-test and an MMD test to measure the distance. We set the significance level $\alpha = 0.05$.

\begin{figure}
    \centering
    \includegraphics[width=\linewidth]{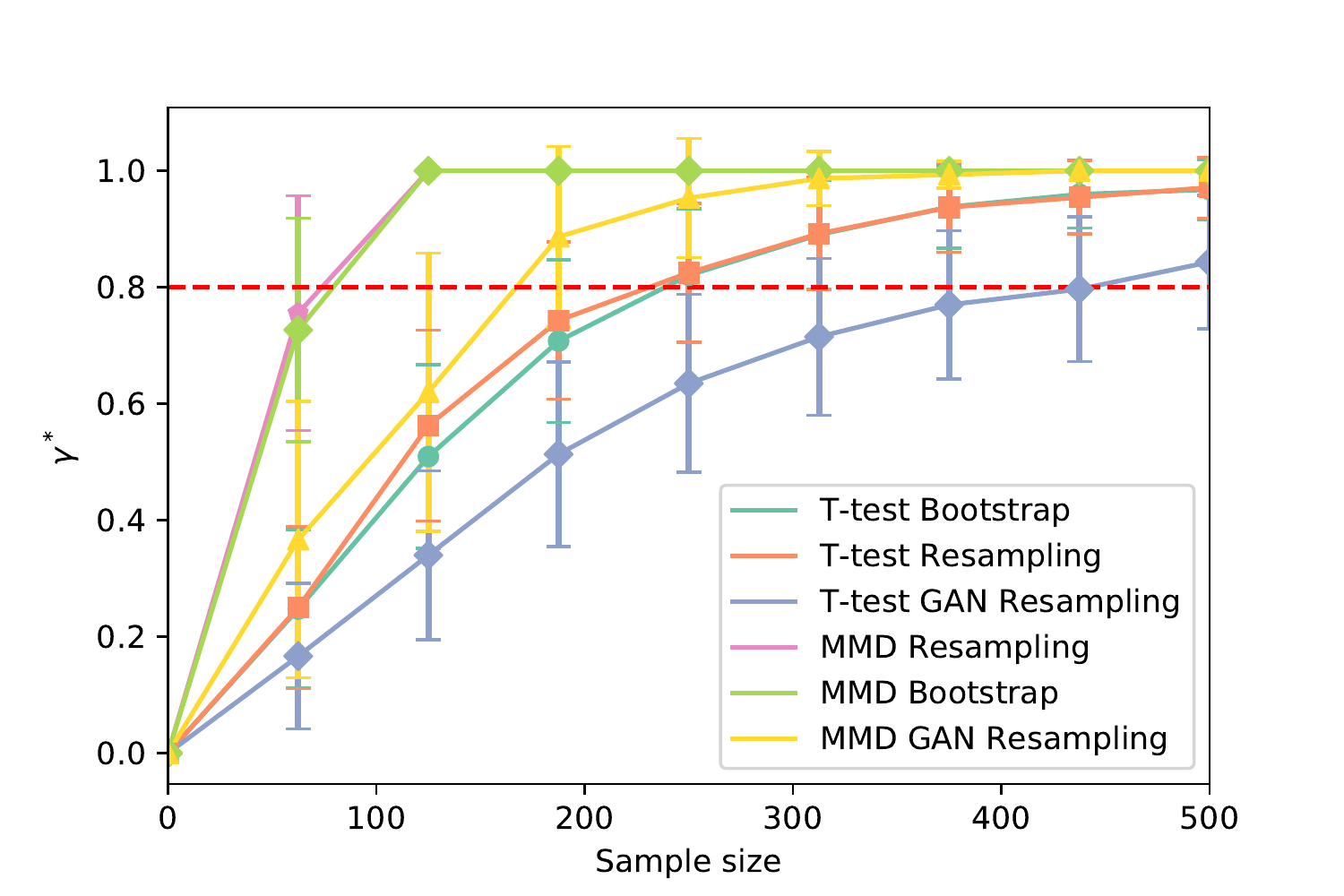}
    \vspace{-0.6cm}
    \caption{Power curves on simulated data. The two multivariate Gaussian distributions are $D_1 \sim \mathcal{N} (\mu = \textbf{0}, \Sigma = \mathbb{I})$  and $D_2 \sim \mathcal{N} (\mu = \textbf{0.3}, \Sigma = \mathbb{I})$ with the dimension of 10. The red line refers to $\gamma^* = 0.8$.}
    \label{fig:sim_power}
\vspace{-0.5cm}
\end{figure}

 Figure~\ref{fig:sim_power} summarizes the results of the t-test and MMD test power curves which compare the two Gaussian distributions (the shaded region illustrates the variances). 
 We briefly outline some observations from the plots in the following. 
 
 i) Using the ``Resampling" strategy outperfoms the ``Bootstrap" as expected. Recall that at each iteration, ``Resampling" obtains samples from the real distributions, which one expects will result in a more powerful test.
 
ii) In most cases, the power curves of synthetic brain data are lower than the curves of real brain data, suggesting that the test is appropriately conservative in many cases. 

\subsection{Synthetic power on fMRI data}
\label{exp:fmri}
Neurovault\cite{gorgolewski2015neurovault} is currently the largest open database of preprocessed neuroimaging data. We evaluate the performance of the proposed synthetic power analyses on the the largest Neurovault functional brain image collection 1952\footnote{The collection is available here: https://neurovault.org/collections/1952}. Collection 1952 is a test-dependent fMRI dataset with 19 meta-labels such as `visual,' `language,' and `calculate.'. The dimensions of the brain images in  collection 1952 are $53 \times 63 \times 46$.
 Following standard preprocessing procedures, a brain image is normalized using the nilearn python package\footnote{http://nilearn.github.io}. 
We first project the data to 10 principle components of fMRI volumes obtained from Principal Component Analysis (PCA) for all the power analysis~\cite{pearson1901liii}. We conduct experiments for two labels having the largest number of fMRI volumes in collection 1952, i.e., ``visual" and ``auditory". We train an ICW-GAN for each label. As an example, we divide collection 1952 into two sets: with the tag ``visual" and without ``visual". We train an ICW-GAN for the two contrast labels of data. We conduct the same procedure for the label ``auditory". Note that there could be an overlap across different labels, e.g., an fMRI data can have both tags ``visual" and ``auditory".

Before the quantitative analysis, we visualize 2D projections of several real and synthetic brain volumes with different tags in Fig.~\ref{fig:syn}. We show real fMRI data in Column (a) and synthetic images in Column (b). Each row represents images with a certain tags, i.e., ``visual" (first row) vs ``non-visual"(second row), and ``auditory" (third row) vs ``non-auditory"(forth row). 
Previous work has examined the quality of synthetic images for data augmentation~\cite{zhuang2018hallucinating}. Specifically, classifiers are trained on the fMRI data with the two contrast tags, and the results indicated that adding synthetic brain images to the original real training dataset helped to improve the performance of classification. All test parameters, e.g., $\alpha$ and the number of trials are the same as for the simulated data.

\begin{figure}[t]
    \centering
    \includegraphics[width=\linewidth]{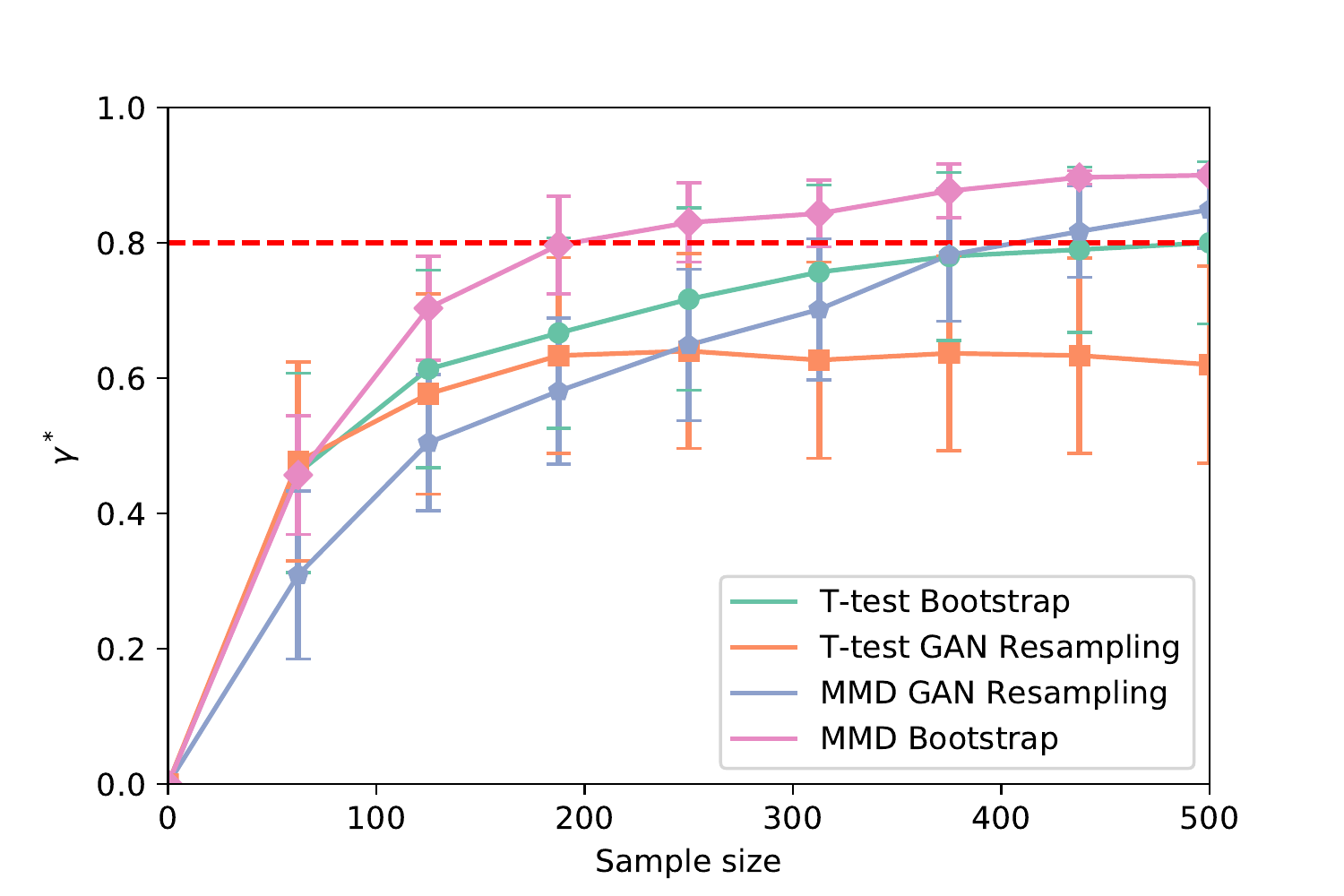}
    \vspace{-0.6cm}
    \caption{Power curves for fMRI data with/without the tag ``visual". The red line refer to $\gamma^* = 0.8$, typically used by for neuroimaging experiments.}
    \label{fig:visual_power}
\vspace{-0.5cm}
\end{figure}
\begin{figure}[t]
    \centering
    \includegraphics[width=\linewidth]{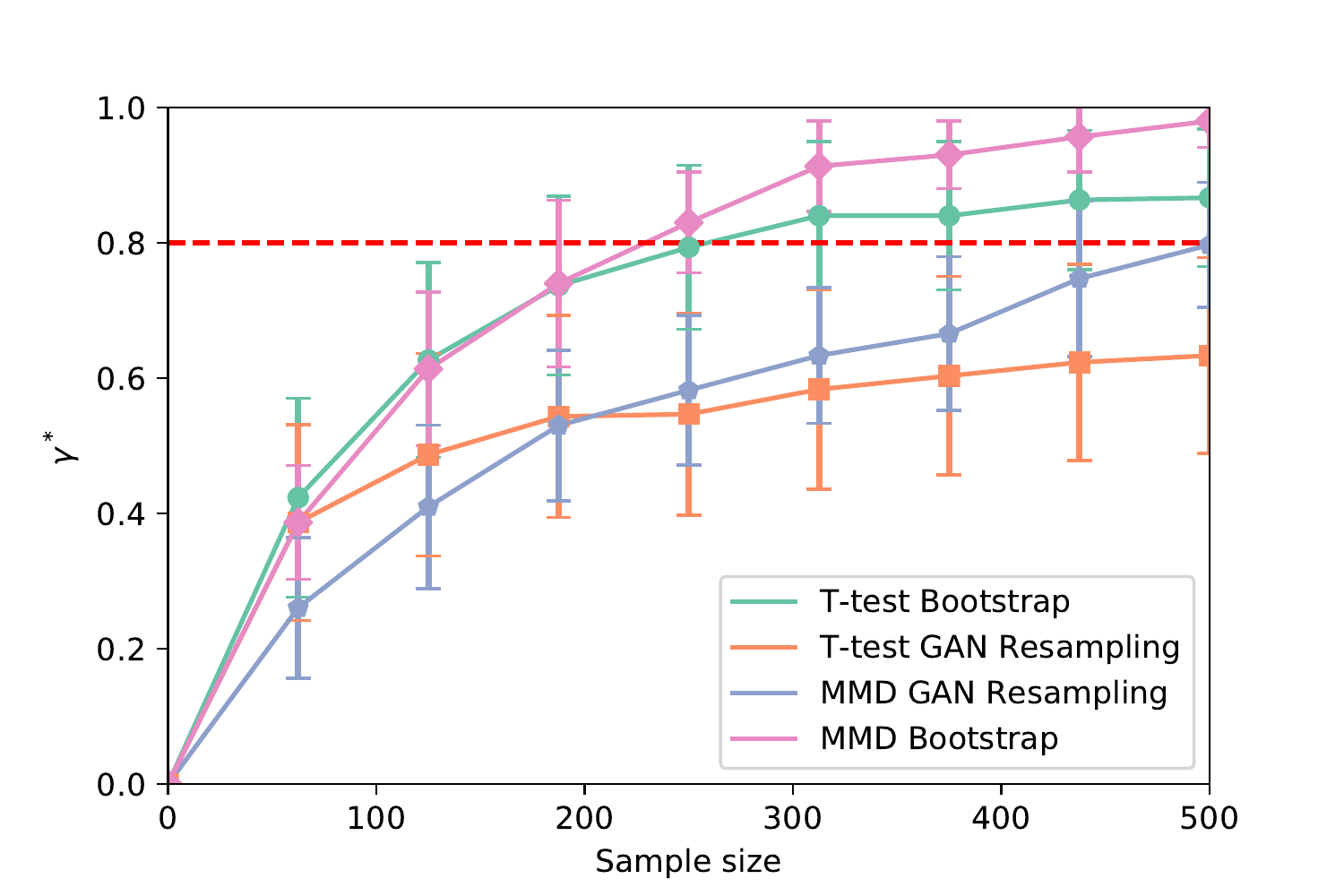}
    \vspace{-0.6cm}
    \caption{Power curves on fMRI data with/without the tag ``auditory". The red line refers to $\gamma^* = 0.8$, typically used for neuroimaging experiments.}
    \label{fig:audi_power}
\vspace{-0.5cm}
\end{figure}

We present the power curves (curves are independently smoothed) for ``visual" and ``auditory" tags in Fig.~\ref{fig:visual_power} and Fig.~\ref{fig:audi_power}.
From the two figures, we notice that the ``MMD bootstrap" curves are the highest performing, achieving 80\% power with fewer samples. The power curves of synthetic data are consistently lower than those of real brain data on the same test, suggesting that the power estimates are conservative (conservative power estimates ideal for sample size selection). Our results recover the empirical observation from \cite{mmd2} that the MMD test outperforms the t-test on data that does not satisfy Gaussian distributional assumptions. For our synthetic brain images, the MMD outperforms the t-test with sample sizes greater than 200. The curves of the real images illustrate similar results. 


\vspace{-0.3cm}
\section{Conclusion}
\label{sec:con}
This work demonstrates the potential utility of synthetic power analyses through experimental evaluation of simulated data and fMRI data. Specifically, we compare the power analysis results of synthetic and real data and illustrate that the GAN model sufficiently recovers the distributional characteristics of the real data with respect to power estimates, which, in turn, translates to accurate sample size estimates. We employ two statistical testing methods, i.e., the t-test and the MMD test, and compare the power in different cases. For imaging data, our results focus on the 3D Conditional Improved Wasserstein Generative Adversarial Networks to synthesize diverse, high-quality brain images. Our empirical results suggest that power analysis results calculated with synthetic data are conservative compared to real data, indicating that the synthetic powers could serve as a low-cost and statistically reliable replacement for real data used in power analyses. 

Synthetic power analyses have distinct advantages over traditional methods -- the simulated data are not restricted by sample size and can synthesize images corresponding to flexible experimental conditions (by conditioning the generative model). For future work, we plan on a more thorough evaluation of synthetic power analysis, including evaluation in real experimental conditions. We are also interested in theoretical analyses of the properties of synthetic power analyses in different settings.
\vspace{-0.3cm}
\bibliographystyle{IEEEtran}  
\bibliography{IEEEabrv,IEEEexample}


\end{document}